\documentclass[10pt,twocolumn,letterpaper]{article}

\usepackage[pagenumbers]{cvpr} 

\usepackage{graphicx}
\usepackage{amsmath}
\usepackage{amssymb}
\usepackage{booktabs}
\usepackage{makecell} 
\usepackage{multirow}
\usepackage{array}
\newcolumntype{C}[1]{>{\centering\let\newline\\\arraybackslash\hspace{0pt}}m{#1}}
\newcolumntype{?}{!{\vrule width 1pt}}
\usepackage[table]{xcolor}

\usepackage[pagebackref,breaklinks,colorlinks]{hyperref}

\usepackage[capitalize]{cleveref}
\crefname{section}{Sec.}{Secs.}
\Crefname{section}{Section}{Sections}
\Crefname{table}{Table}{Tables}
\crefname{table}{Tab.}{Tabs.}

\def\cvprPaperID{*****} 
\def\confName{CVPR}
\def\confYear{2023}

\begin{document}

\title{PIDNet: A Real-time Semantic Segmentation Network Inspired by PID Controllers}

\author{Jiacong Xu \qquad
    Zixiang Xiong  \qquad
    Shankar P. Bhattacharyya \\
    Dept of ECE, Texas A\&M University, College Station, TX 77843 \\
{\tt\small jxu155@jhu.edu, zx@ece.tamu.edu, spb@tamu.edu}
}
\maketitle

\begin{abstract}
Two-branch network architecture has shown its efficiency and effectiveness in real-time semantic segmentation tasks. However, direct fusion of high-resolution details and low-frequency context has the drawback of detailed features being easily overwhelmed by surrounding contextual information. This overshoot phenomenon limits the improvement of the segmentation accuracy of existing two-branch models. In this paper, we make a connection between Convolutional Neural Networks (CNN) and Proportional-Integral-Derivative (PID) controllers and reveal that a two-branch network is equivalent to a Proportional-Integral (PI) controller, which inherently suffers from similar overshoot issues. To alleviate this problem, we propose a novel three-branch network architecture: PIDNet, which contains three branches to parse detailed, context and boundary information, respectively, and employs boundary attention to guide the fusion of detailed and context branches. Our family of PIDNets achieve the best trade-off between inference speed and accuracy and their accuracy surpasses all the existing models with similar inference speed on the Cityscapes and CamVid datasets. Specifically, PIDNet-S achieves 78.6\% mIOU with inference speed of 93.2 FPS on Cityscapes and 80.1\% mIOU with speed of 153.7 FPS on CamVid.
\end{abstract}
\footnotetext[1]{Work supported in part by NSF grants ECCS-1923803 and CCF-2007527.}

\section{Introduction}
\label{sec:intro}
\begin{figure}
\centering
    \includegraphics[width=0.475\textwidth]{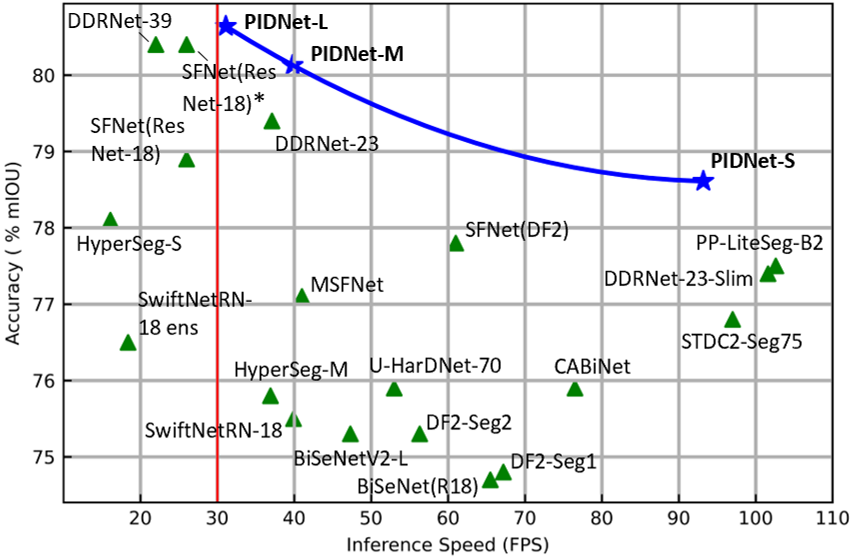}
\caption{The trade-off between inference speed and accuracy (reported) for real-time models on the Cityscapes \cite{cityscapes} test set. Blue stars refer to our models while green triangles represent others.}
\label{fig:score}
\end{figure}

Proportional-Integral-Derivative (PID) Controller is a classic concept that has been widely applied in modern dynamic systems and processes such as robotic manipulation \cite{pid_robot}, chemical processes \cite{pid_chemical}, and power systems \cite{pid_power}. Even though many advanced control strategies with better control performance have been developed in recent years, PID controller is still the go-to choice for most industry applications due to its simplicity and robustness. Furthermore, the idea of PID controller has been extended to many other areas. For example, researchers introduced the PID concept to image denoising \cite{pid_image}, stochastic gradient decent \cite{pid_stochastic} and numerical optimization \cite{pid_pso} for better algorithm performance. In this paper, we devise a novel architecture for real-time semantic segmentation tasks by employing the basic concept of PID controller and demonstrate that the performance of our model surpasses all the previous works and achieves the best trade-off between inference speed and accuracy, as illustrated in Figure \ref{fig:score}, by extensive experiments.

Semantic segmentation is a fundamental task for visual scene parsing with the objective of assigning each pixel in the input image to a specific class label. 
With the increasing demand of intelligence, semantic segmentation has become the basic perception component for applications such as autonomous driving \cite{ss_auto}, medical imaging diagnosis \cite{ss_medical} and remote sensing imagery\cite{ss_remote}. 
Starting from FCN \cite{fcn}, which achieved great improvement over traditional methods, deep convnets gradually dominated the semantic segmentation field and many representative models have been proposed \cite{unet, segnet, deeplab, pspnet,hrnet,transformer}. 
For better performance, various strategies were introduced to equip these models with the capability of learning contextual dependencies among pixels in large scale without missing important details. 
Even though these models achieve encouraging segmentation accuracy, too much computational cost are required, which significantly hinder their application in real-time scenarios, such as autonomous vehicle \cite{ss_auto} and robot surgery \cite{surgery}.

To meet real-time or mobile requirements, researchers have come up with many efficient and effective models in the past for semantic segmentation. Specifically, ENet \cite{enet} adopted lightweight decoder and downsampled the feature maps in early stages. ICNet \cite{icnet} encoded small-size inputs in complex and deep path to parse the high-level semantics. MobileNets \cite{mobilenets, mobilenetv2} replaced traditional convolutions with depth-wise separable convolutions. These early works reduced the latency and memory usage of segmentation models, but low accuracy significantly limits their real-world application. Recently, many novel and promising models based on Two-Branch Network (TBN) architecture have been proposed in the literature and achieve SOTA trade-off between speed and accuracy \cite{bisenet,fast_scnn, contextnet, stdc, ddrnet}. 

\begin{figure}
\centering
    \includegraphics[width=0.45\textwidth]{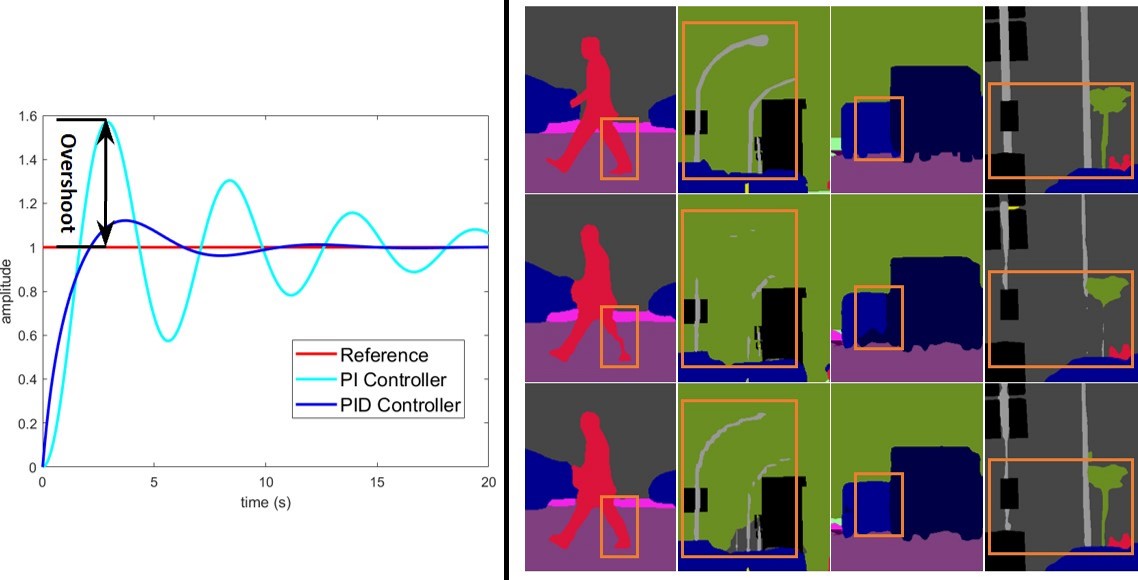}
\caption{Overshoot issue for dynamic system (left $|$) and image segmentation ($|$ right). Left $|$: Step responses of PI and PID controllers for a second-order system; $|$ Right: From the first row to the last row, the images are cropped from ground truth, outputs of DDRNet-23 \cite{ddrnet} and ADB-Bag-DDRNet-23 (ours), respectively.}
\label{fig:overshoot}
\end{figure}

In this paper, we view the architecture of TBNs from the prospective of PID controller and point out that a TBN is equivalent to a PI controller, which suffers from the overshoot issue as illustrated in Figure \ref{fig:overshoot}. 
To alleviate this problem, we devise a novel three-branch network architecture, namely PIDNet, and demonstrate its superiority on Cityscapes \cite{cityscapes}, CamVid \cite{camvid} and PASCAL Context \cite{pascal_context} datasets. We also provide ablation study and feature visualization for better understanding of the functionality of each module in PIDNet. The source code can be accessed via: \url{https://github.com/XuJiacong/PIDNet}

The main contributions of this paper are three-fold:
\begin{itemize}
  \item We make a connection between deep CNN and PID controller and propose a family of three-branch networks based on the PID controller architecture. 
  \item Efficient modules, such as Bag fusion module designed to balance detailed and context features, are proposed to boost the performance of PIDNets. 
  \item PIDNet achieves the best trade-off between inference speed and accuracy among all the existing models. In particular, PIDNet-S achieves $78.6\%$ mIOU with speed of $93.2$ FPS and PIDNet-L presents the highest accuracy ($80.6\%$ mIOU) in real-time doman on Cityscapes test set without acceleration tools.
\end{itemize}

\section{Related Work}
\label{sec:related}
Representative methods towards high-accuracy and real-time requirements are discussed separately in this section.

\subsection{High-accuracy Semantic Segmentation}
Early approaches for semantic segmentation were based on an encoder-decoder architecture \cite{fcn, unet, segnet}, where the encoder gradually enlarges its receptive field by strided convolutions or pooling operations and the decoder recovers detailed information from high-level semantics using deconvolutions or upsampling. 
However, spatial details could be easily ignored in the process of downsampling for encoder-decoder network. To alleviate this problem, dilated convolution \cite{dilated} was proposed to enlarge the field of view without reducing the spatial resolution. Based on this, DeepLab series \cite{deeplabv2,deeplabv3,deeplabv3+} achieved great improvement over previous works by employing dilated convolution with different dilation rates in the network. Note that dilated convolution is not suitable for hardware implementation due to its non-contiguous memory accesses. 
PSPNet \cite{pspnet} introduced a pyramid pooling module (PPM) to parse multi-scale context information and HRNet \cite{hrnet} utilized multiple paths and bilateral connections to learn and fuse the representations in different scales. Inspired from the long-range dependency parsing ability of attention mechanism \cite{attention} for language machine, non-local operation \cite{nonlocal} was introduced into computer vision and led to many accurate models \cite{danet, ccnet, ocr}. 

\subsection{Real-time Semantic Segmentation}
Many network architectures have been proposed to achieve the best trade-off between inference speed and accuracy, which could be roughly summarized as below. 

{\bf Light-weight encoder and decoder}
SwiftNet \cite{swiftnet} employed one low-resolution input to obtain high-level semantics and another high-resolution input to provide sufficient details for its lightweight decoder. DFANet \cite{dfanet} introduced a light-weight backbone by modifying the architecture of Xception \cite{xception}, which was based on depth-wise separable convolution, and reduced the input size for faster inference speed. 
ShuffleSeg \cite{shuffleseg} adopted ShuffleNet \cite{shufflenet}, which combined channel shuffling and group convolution, as its backbone to reduce the computational cost. However, most of these networks are still in the form of encoder-decoder architecture and they require the information flow go through the deep encoder and then reverse back to pass the decoder, which introduces too much latency. Besides, since the optimization for depth-wise separable convolution on GPU is not mature, traditional convolution presents faster speed while having more FLOPs and parameters \cite{swiftnet}. Thus, we seek for more efficient model that avoids convolution factorization and encoder-decoder architecture.

{\bf Two-branch network architecture}
Contextual dependency can be extracted by large receptive field, and spatial details are vital for boundary delineation and small-scale object recognition. To take both sides into account, authors of BiSeNet \cite{bisenet} proposed a two-branch network (TBN) architecture, which contains two branches with different depths for context embedding and detail parsing along with a feature fusion module (FFM) to fuse the context and detailed information. 
Several follow-up works based on this architecture have been proposed to boost its representation ability or reduce its model complexity \cite{fast_scnn, contextnet, bisenetv2}. Specifically, DDRNet \cite{ddrnet} introduced bilateral connections to enhance information exchange between context and detailed branches, achieving state-of-the-art results in real-time semantic segmentation. Nevertheless, direct fusion of original detailed semantics and low-frequency context information has the risk of that object boundaries being overly corroded by surrounding pixels and small objects being overwhelmed by adjacent large ones (as shown in Figure \ref{fig:overshoot} and \ref{fig:pid_ana}). 


\section{Method}
\label{sec:method}

\begin{figure}
\centering
    \includegraphics[width=0.475\textwidth]{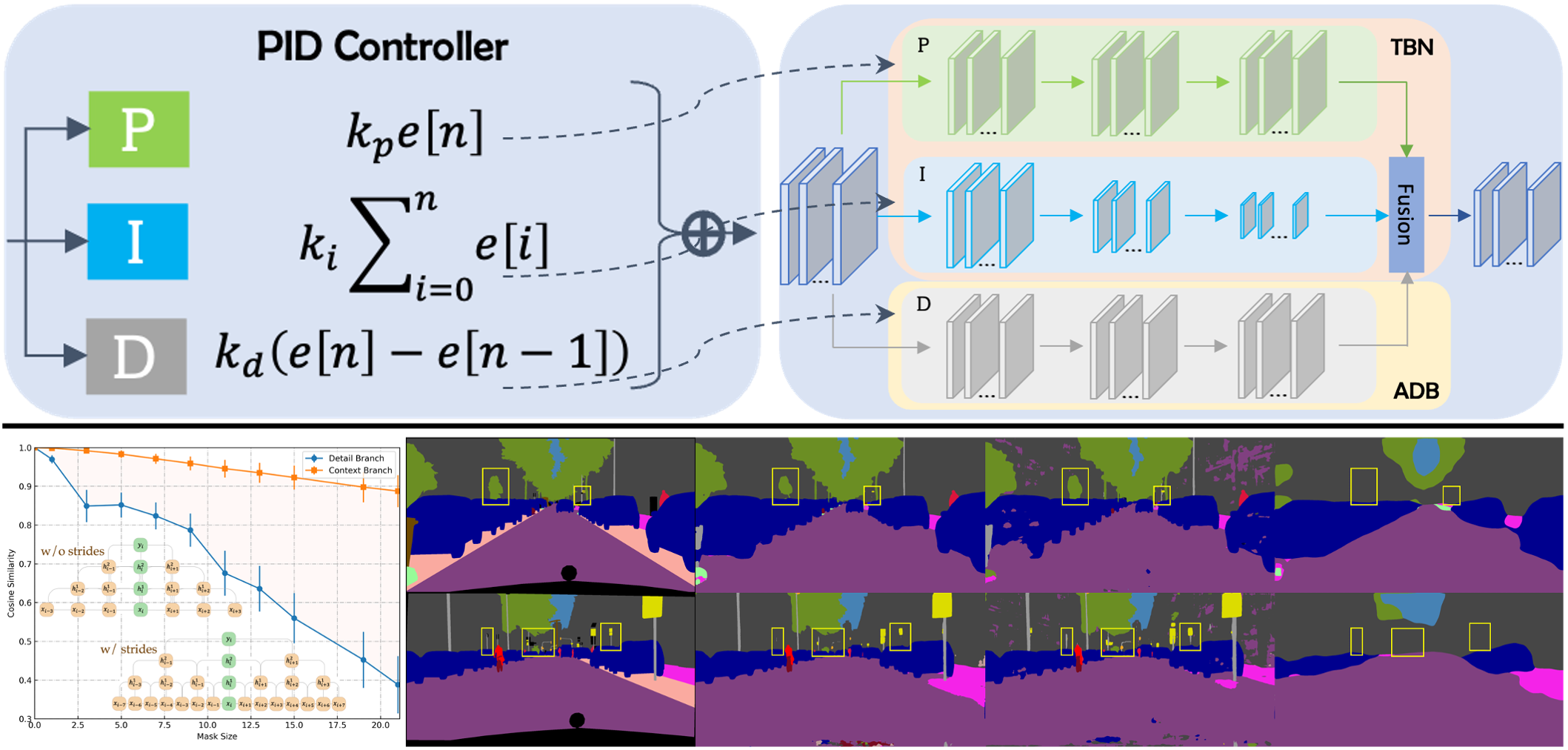}
\caption{Upper $|$: The analogy between PID controller and proposed network; $|$ Bottom: Left: Zero out surrounding mask area and calculate the similarity between current and original features for each pixel; Right: From the first to the last column, the images refer to ground truth, predictions of all branches, the detailed branch only, and the context branch only of DDRNet-23.}
\label{fig:pid_ana}
\end{figure}

A PID controller contains three components: a proportional (P) controller, an integral (I) controller and a derivative (D) controller, as illustrated in Figure \ref{fig:pid_ana}-Upper. The implementation of PI controller could be written as:
\begin{equation}
\label{pi}
c_{out}[n]=k_{p}e[n]+k_{i}\sum^{n}_{i=0}e[i]
\end{equation}
P controller focuses on current signal, while I controller accumulates all the past signals. Due to the inertia effect of accumulation, overshoot will happen to the output of simple PI controller when the signal changes oppositely. Then, D controller was introduced and if the signal become smaller, the D component will become negative and serves as a damper to reduce the overshoot. Similarly, TBNs parse the context and detailed information by multiple convolutional layers with and without strides, respectively. Consider a simple 1D example, where both detailed and context branches consist of 3 layers without BNs and ReLUs. Then, the output maps can be calculated as:
\begin{equation}
\label{conv3_d}
O_{D}[i]=K^{D}_{i-3}I[i-3]+...+K^{D}_{i}I[i]+...+K^{D}_{i+3}I[i+3]
\end{equation}
\begin{equation}
\label{conv3_c}
O_{C}[i]=K^{C}_{i-7}I[i-7]+...+K^{C}_{i}I[i]+...+K^{C}_{i+7}I[i+7]
\end{equation}
where, $K^{D}_{i}=k_{31}k_{22}k_{13}+k_{31}k_{23}k_{12}+k_{32}k_{21}k_{13}+k_{32}k_{22}k_{12}+k_{32}k_{23}k_{13}+k_{33}k_{21}k_{12}+k_{33}k_{22}k_{11}$ and $K^{C}_{i}=k_{32}k_{22}k_{12}$. Here, $k_{mn}$ refers to the $n$-th value of the kernel in layer $m$. Since $|k_{mn}|$ are mostly distributed in $(0, 0.01)$ (92\% for DDRNet-23) and are bounded by 1, the coefficient for each item will decrease exponentially with more layers. Thus, for each input vector, a larger number of items means a higher possibility to contribute to the final output. For detail branch, $I[i-1]$, $I[i]$, and $I[i+1]$ occupy over 70\% of the total items, which means that \textbf{the detail branch focuses more on the local information}. On the contrary, $I[i-1]$, $I[i]$, and $I[i+1]$ only occupies less than 26\% of the total items in context branch, so \textbf{the context branch emphasizes the surrounding information}. Figure \ref{fig:pid_ana}-Bottom shows that the context branch is less sensitive to the change of local information than the detail branch. The behavior of detail and context branches in the spatial domain is similar to the P (current) and I (all previous) controllers in time domain.

\begin{figure*}
\centering
    \includegraphics[scale=0.93]{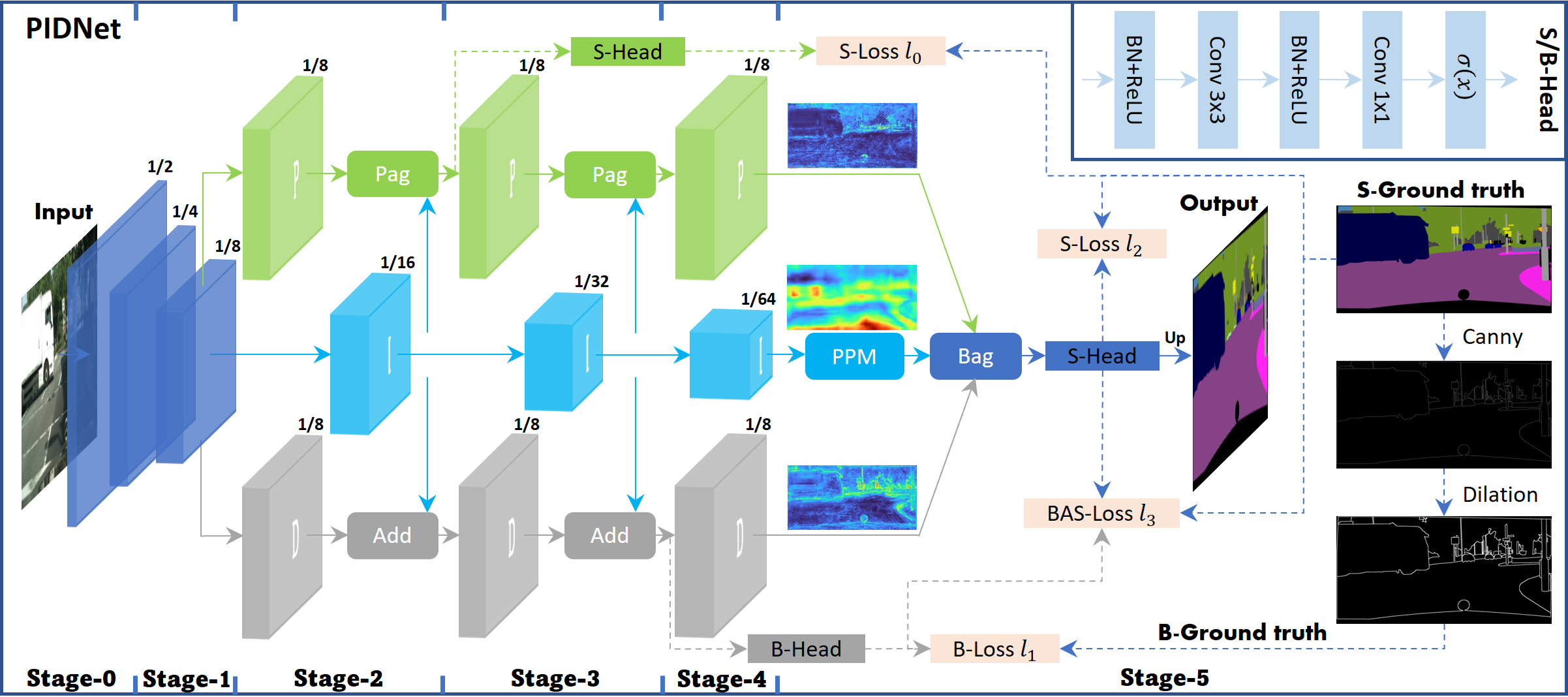}
\caption{\textbf{An overview of the basic architecture of our proposed Proportional-Integral-Derivative Network (PIDNet).} S and B denote semantic and boundary, and Add and Up refer to element-wise summation and bilinear Upsampling operation, respectively; BAS-Loss represents the boundary-awareness CE loss \cite{boundary-aware}. Dashed lines and associate blocks will be ignored in the inference stage.}
\label{fig:pidnet}
\end{figure*}

Replace $z^{-1}$ by $e^{-j\omega}$ in the z-transform of a PID controller, which could be represented as:
\begin{equation}
\label{pid_z}
C(z)=k_p+k_i(1-e^{-j\omega})^{-1}+k_d(1-e^{-j\omega})
\end{equation} 
when the input frequency $\omega$ increases, the gain of I and D controllers will becomes smaller and larger, respectively, so the P, I, and D controllers work as allpass, lowpass filter, and highpass filter. Since PI controller focuses more on the low-frequency part of the input signal and cannot react immediately to the rapid change of the signal, it inherently suffers from the overshoot problem. The D controller reduces the overshoot by enabling the control output sensitive to the change of input signal. Figure \ref{fig:pid_ana}-Bottom shows that the detail branch parses all kinds of semantic information even though not accurate, whereas \textbf{the context branch aggregates the low-frequency context information} and works similarly with a large averaging filter on semantics. Direct fusion of detailed and context information leads to missing of some detailed features. Thus, we conclude that TBN is equivalent to a PI controller in Fourier domain.

\subsection{PIDNet: A Novel Three-branch Network} 
To mitigate the overshoot problem, we attach an auxiliary derivative branch (ADB) to the TBN to mimic the PID controller spatially and highlight the high-frequency semantic information. 
The semantics for pixels inside each object are consistent and only become inconsistent along the boundary of adjacent objects, so the difference of semantics is nonzero only at the object boundary and the objective of ADB is boundary detection. Accordingly, we establish a new three-branch real-time semantic segmentation architecture, namely Proportional-Integral-Derivative Network (PIDNet), which is shown in Figure \ref{fig:pidnet}. 

PIDNet possesses three branches with complementary responsibilities: the proportional (P) branch parses and preserves detailed information in high-resolution feature maps; the integral (I) branch aggregates context information both locally and globally to parse long-range dependencies; and the derivative (D) branch extracts high-frequency features to predict boundary regions. 
As \cite{ddrnet}, we also adopt cascaded residual blocks \cite{resnet} as the backbone for hardware friendliness. 
Besides, the depths for the P, I and D branches are set to be moderate, deep and shallow for efficient implementation. Consequently, a family of PIDNets (PIDNet-S, M and L) are generated by deepening and widening the model.

Following \cite{bisenetv2, ddrnet, sfnet}, we place a semantic head at the output of the first Pag module to generate the extra semantic loss $l_{0}$ for better optimization of entire network. Instead of dice loss \cite{crispy}, weighted binary cross entropy loss  $l_{1}$ is adopted to deal with the imbalanced problem of boundary detection since coarse boundary is preferred to highlight the boundary region and enhance the features for small objects. $l_{2}$ and $l_{3}$ represents the CE loss, while we utilize the boundary-awareness CE loss \cite{boundary-aware} for $l_{3}$ using the output of boundary head to coordinate semantic segmentation and boundary detection tasks and enhance the function of Bag module. The calculation of BAS-Loss can be written as:
\begin{equation}
\label{bas}
l_{3}=-\sum_{i,c}\{1: b_{i} > t\}(s_{i,c}log\hat{s_{i,c}})
\end{equation}
where t refers to predefined threshold and $b_{i}$, $s_{i,c}$ and $\hat{s_{i,c}}$ are the output of boundary head, segmentation ground-truth and prediction result of the i-th pixel for class c, respectively. Therefore, the final loss for PIDNet is:
\begin{equation}
\label{loss}
Loss = \lambda_{0}l_{0} + \lambda_{1}l_{1} + \lambda_{2}l_{2} + \lambda_{3}l_{3}
\end{equation}
Empirically, we set the parameters for the training loss of PIDNet as $\lambda_{0}=0.4$, $\lambda_{1}=20$, $\lambda_{2}=1$, $\lambda_{3}=1$ and $t=0.8$.

\subsection{Pag: Learning High-level Semantics Selectively}
\begin{figure}[h]
\centering
    \includegraphics[width=0.47\textwidth]{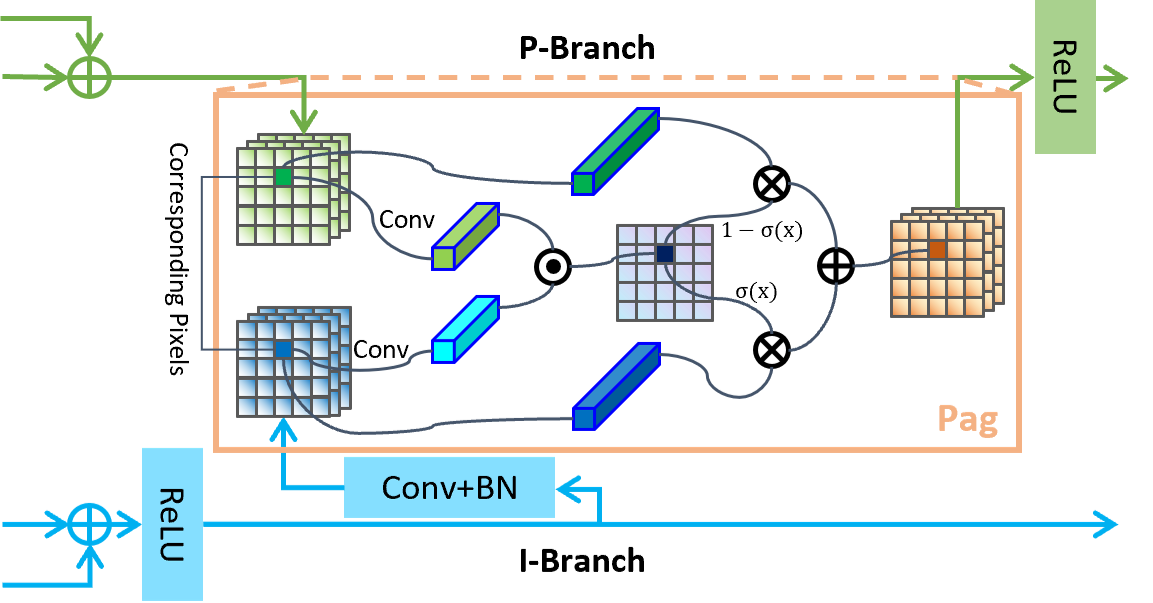}
\caption{Illustration of Pag module. $\sigma(x)$ denotes the Sigmoid function; The kernel sizes of all the convolutions here are  $1\times1$.}
\label{fig:pag}
\end{figure}

The lateral connection utilized in \cite{swiftnet, ddrnet, hrnet} enhances the information transmission between feature maps in different scales and improves the representation ability of their models. In PIDNet, the rich and accurate semantic information provided by I branch is crucial for detail parsing and boundary detection of the P and D branches, both of which contain relatively less layers and channels. Thus, we treat the I branch as the backup for other two branches and enable it to provide required information to them. Different from the D branch that directly adds the provided feature maps, we introduce a \textbf{P}ixel-\textbf{a}ttention-\textbf{g}uided fusion module (Pag), which is shown in Figure \ref{fig:pag}, 
for the P branch to selectively learn the useful semantic features from I branch without being overwhelmed.
The underlying concept for Pag is borrowed from attention mechanisms \cite{attention}. Define the vectors for the corresponding pixels in feature maps from the P and I branch as $\Vec{v_{p}}$ and $\Vec{v_{i}}$, respectively, then the output of the Sigmoid function could be represented as:
\begin{equation}
\label{sig_pag}
\sigma=Sigmoid(f_{p}(\Vec{v_{p}})\cdot f_{i}(\Vec{v_{i}}))
\end{equation}
where $\sigma$ indicates the possibility of these two pixels belonging to the same object. If $\sigma$ is high, we trust $\Vec{v_{i}}$ more since the I branch is semantically rich and accurate, and vise versa. Thus, the output of the Pag can be written as:
\begin{equation}
\label{pag}
Out_{Pag}=\sigma \Vec{v_{i}} + (1-\sigma)\Vec{v_{p}}
\end{equation}

\subsection{PAPPM: Fast Aggregation of Contexts}
For better global scene prior construction, PSPNet \cite{pspnet} introduced a pyramid pooling module (PPM), which concatenates multi-scale pooling maps before convolution layer to form local and global context representations. Deep Aggregation PPM (DAPPM) proposed by \cite{ddrnet} further improved the context embedding ability of PPM and showed superior performance. 
\begin{figure}[h]
\centering
    \includegraphics[width=0.46\textwidth]{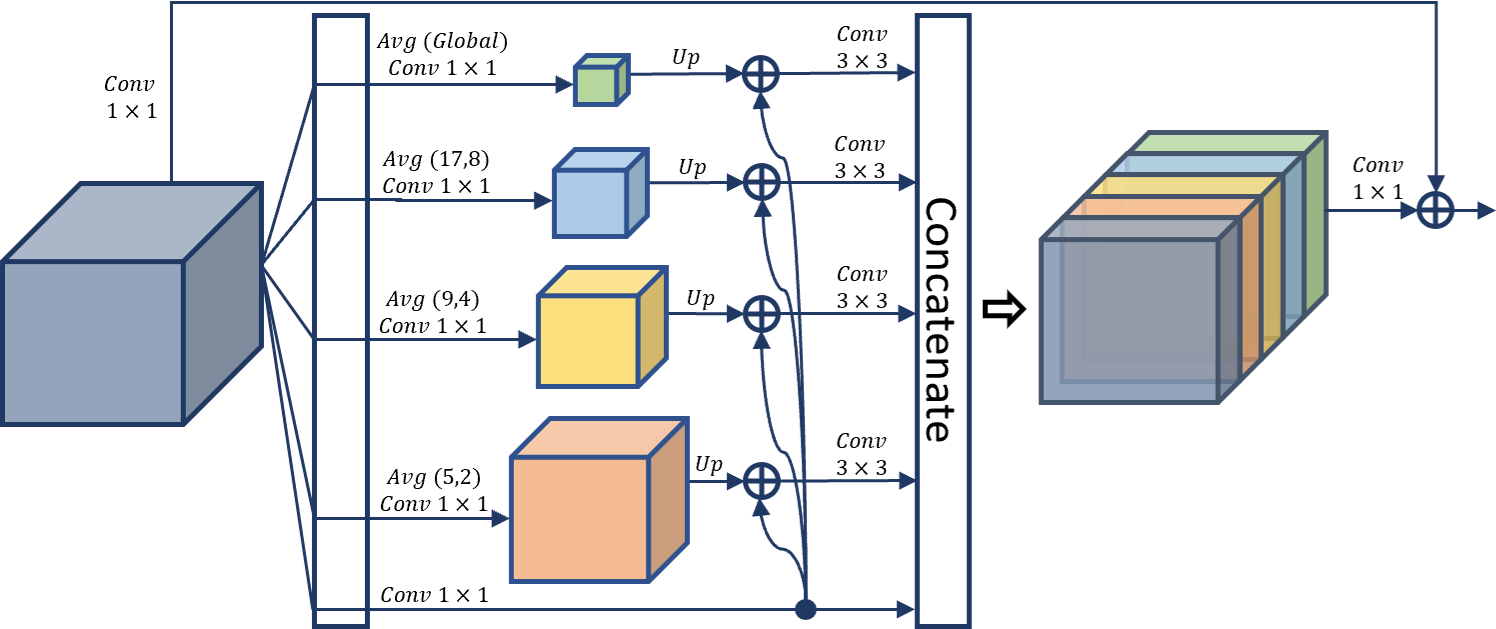}
\caption{The parallel structure of PAPPM. Avg (5,2) means average pooling with kernel size of 5$\times$5 and strides of 2.}
\label{fig:pappm}
\end{figure}
Nevertheless, the computation process of DAPPM cannot be parallelized regarding its depth, which is time-consuming and DAPPM contains too many channels for each scale, which may surpasses the representation ability of lightweight models. 
Thus, we modify the connections in DAPPM to make it parallelizable, which is shown in Figure \ref{fig:pappm}, and reduce the number of channels for each scale from 128 to 96. This new context harvesting module is called Parallel Aggregation PPM (PAPPM) and is applied in PIDNet-M and PIDNet-S to guarantee their speeds. For our deep model: PIDNet-L, we still choose the DAPPM considering its depth but reduce its number of channels for less computation and faster speed.

\subsection{Bag: Balancing the Details and Contexts} 
\begin{figure}[ht]
\centering
    \includegraphics[width=0.46\textwidth]{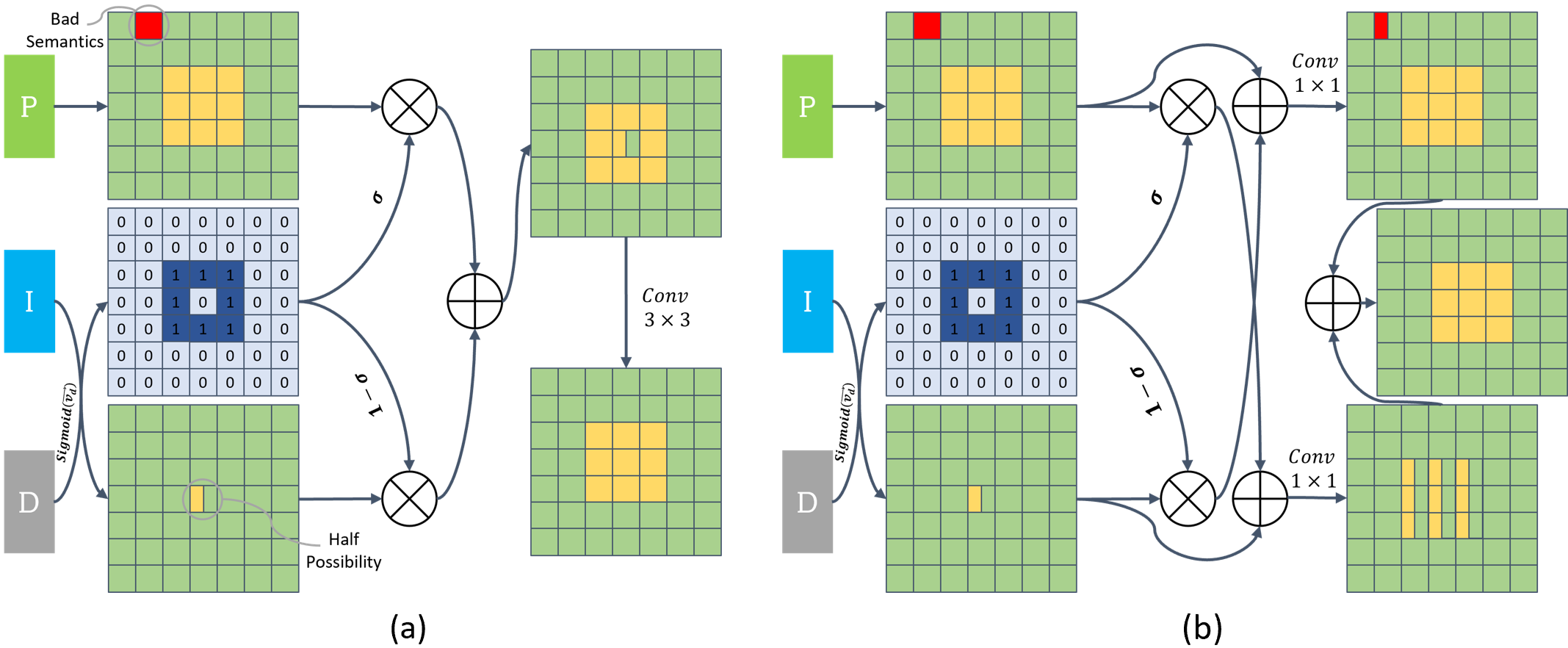}
\caption{Single channel implementations of (a) Bag and (b) Light-Bag modules in extreme case. P, I and D refer to the outputs of detailed, context and boundary branches, respectively. $\sigma$ denotes the output of Sigmoid function.}
\label{fig:bag}
\end{figure}

Given the boundary features extracted by ADB, we employ boundary attention to guide the fusion of detailed (P) and context (I) representations. Specifically, we design a \textbf{B}oundary-\textbf{a}ttention-\textbf{g}uided fusion module (Bag), shown in Figure \ref{fig:bag}, to fill the high-frequency and low-frequency areas with detailed and context features, respectively. Note that the context branch is semantically accurate but it loses too much spatial and geometric details especially for the boundary region and small object. Thanks to the detailed branch, which preserves spatial details better, we force the model to trust the detailed branch more along the boundary region and utilize the context features to fill other areas.
Define the vectors for the corresponding pixels of P, I and D feature maps as $\Vec{v_{p}}$, $\Vec{v_{i}}$ and $\Vec{v_{d}}$, respectively, then the outputs of Sigmoid , Bag and Light-Bag could be represented as: 
\begin{equation}
\label{sig_bag}
\boldsymbol{\sigma}=Sigmoid(\Vec{v_{d}})
\end{equation}
\begin{equation}
\label{bag}
Out_{bag}=f_{out}((1-\boldsymbol{\sigma})\otimes \Vec{v_{i}}+\boldsymbol{\sigma}\otimes \Vec{v_{p}})
\end{equation}
\begin{equation}
\label{light_bag}
Out_{light}=f_{p}((1-\boldsymbol{\sigma})\otimes \Vec{v_{i}}+\Vec{v_{p}})+f_{i}(\boldsymbol{\sigma}\otimes \Vec{v_{p}}+\Vec{v_{i}})
\end{equation}
where $f$ refers to the composition of convolutions, batch normalizations and ReLUs. Even though we replace $3\times3$ convolutions in Bag by two $1\times1$ convolutions in Light-Bag, the functionalities of Bag and Light-Bag are similar, that is when $\sigma>0.5$ the model trusts more on detailed features, otherwise context information is preferred.

\section{Experiment}
\label{sec:exp}
In this section, our models will be trained and tested on Cityscapes, CamVid and PASCAL Context benchmarks. 

\subsection{Datasets}
\noindent
\textbf{Cityscapes.} Cityscapes \cite{cityscapes} is one of the most well-known urban scene parsing datasets, which contains 5000 images collected from the car perspective in different cities. These images are divided into sets with numbers of 2975, 500, and 1525 for training, validation and testing. The image resolution is 2048$\times$1024, which is challenging for real-time models. Only the fine annotated dataset is used here.

\noindent
\textbf{CamVid.} CamVid \cite{camvid} provides 701 images of driving scenes, which is partitioned into 367, 101 and 233 for training, validation and test. The image resolution is of 960$\times$720 and the number of annotated categories is 32, of which 11 classes are used for fair comparison with previous works.

\noindent
\textbf{PASCAL Context.} Semantic labeling for whole scene is provided in PASCAL Context \cite{pascal_context}, which contains 4998 images for training and 5105 images for validation. While this dataset is mainly used for benchmarking high-accuracy models, we utilized it here to show the generalization ability of PIDNets. Both 59 and 60-class scenarios are evaluated.

\subsection{Implementation Details}
\noindent
\textbf{Pretraining.} Before fine-tuning our models, we pre-train them by ImageNet \cite{imagenet} as most of previous works doing \cite{ddrnet, hyperseg,swiftnet}. We remove the D branch and directly merge the features in final stage to construct the classification models. The total number of training epochs is 90 and the learning rate is scheduled to be 0.1 initially and multiplied by 0.1 at epoch 30 and 60. The images are randomly cropped into 224$\times$224 and flipped horizontally for data augmentation.

\noindent
\textbf{Training.} Our training protocols are almost the same as previous works \cite{bisenet, ddrnet, stdc}. Specifically, we adopt the poly strategy to update the learning rate and random cropping, random horizontal flipping, and random scaling in the range of $[0.5, 2.0]$ for data augmentation. The number of training epochs, the initial learning rate, weight decay, cropped size and batch size for Cityscapes, CamVid and PASCAL Context could be summarized as [484, $1e^{-2}$, $5e^{-4}$, 1024$\times$1024, 12], [200, $1e^{-3}$, $5e^{-4}$, 960$\times$720, 12] and [200, $1e^{-3}$, $1e^{-4}$, 520$\times$520, 16], respectively. Following \cite{bisenetv2, ddrnet}, we fine-tune the Cityscapes pretrained models for CamVid and stop the training process when $lr<5e^{-4}$ to avoid overfitting.

\noindent
\textbf{Inference.} Before testing, our models are trained by both train and val set for Cityscapes and CamVid. We measure the inference speed on the platform consists of single RTX 3090, PyTorch 1.8, CUDA 11.2, cuDNN 8.0 and Windows-Conda environment. Using the measurement protocol proposed by \cite{fasterseg} and following \cite{ddrnet, swiftnet, msfnet}, we integrate the batch normalization into the convolutional layers and set the batch size to be 1 for measurement of inference speed.

\subsection{Ablation Study}
\noindent
\textbf{ADB for Two-branch Networks.} To demonstrate the effectiveness of PID methodology, we combine ADB and Bag with existing models. Here, two representative two-branch networks: BiSeNet \cite{bisenet} and DDRNet \cite{ddrnet} equipped with ADB and Bag are implemented and achieve much higher accuracy on Cityscapes val set compared with their original models, which is shown in Table \ref{tab:adb_bag}. However, additional computation significantly slow down their inference speed, which then triggers us to establish PIDNet.
\begin{table}[ht]
\centering
\begin{tabular}{ccccc} 
\Xhline{1pt}
\multirow{2}{*}{Model}   & \multicolumn{2}{c}{ADB-Bag} & \multirow{2}{*}{mIOU} & \multirow{2}{*}{FPS}  \\ 
\cline{2-3}
                         & w/o & w/                      &                       &                       \\ 
\Xhline{1pt}
\multirow{2}{*}{BiSeNet(Res18)} & \checkmark   &                        & 75.4                  & 63.2                  \\ 
\cline{2-5}
                         &     & \checkmark                      & \textbf{76.7}                  & 52.1                 \\ 
\hline
\multirow{2}{*}{DDRNet-23}  & \checkmark   &                        & 79.5                  & 51.4                  \\ 
\cline{2-5}
                         &     & \checkmark                      & \textbf{80.0}                  & 39.2                  \\
\Xhline{1pt}
\end{tabular}
\caption{Ablation study of ADB-Bag for BiSeNet and DDRNet.}
\label{tab:adb_bag}
\end{table}

\begin{figure}[t]
\centering
    \includegraphics[width=8.3cm]{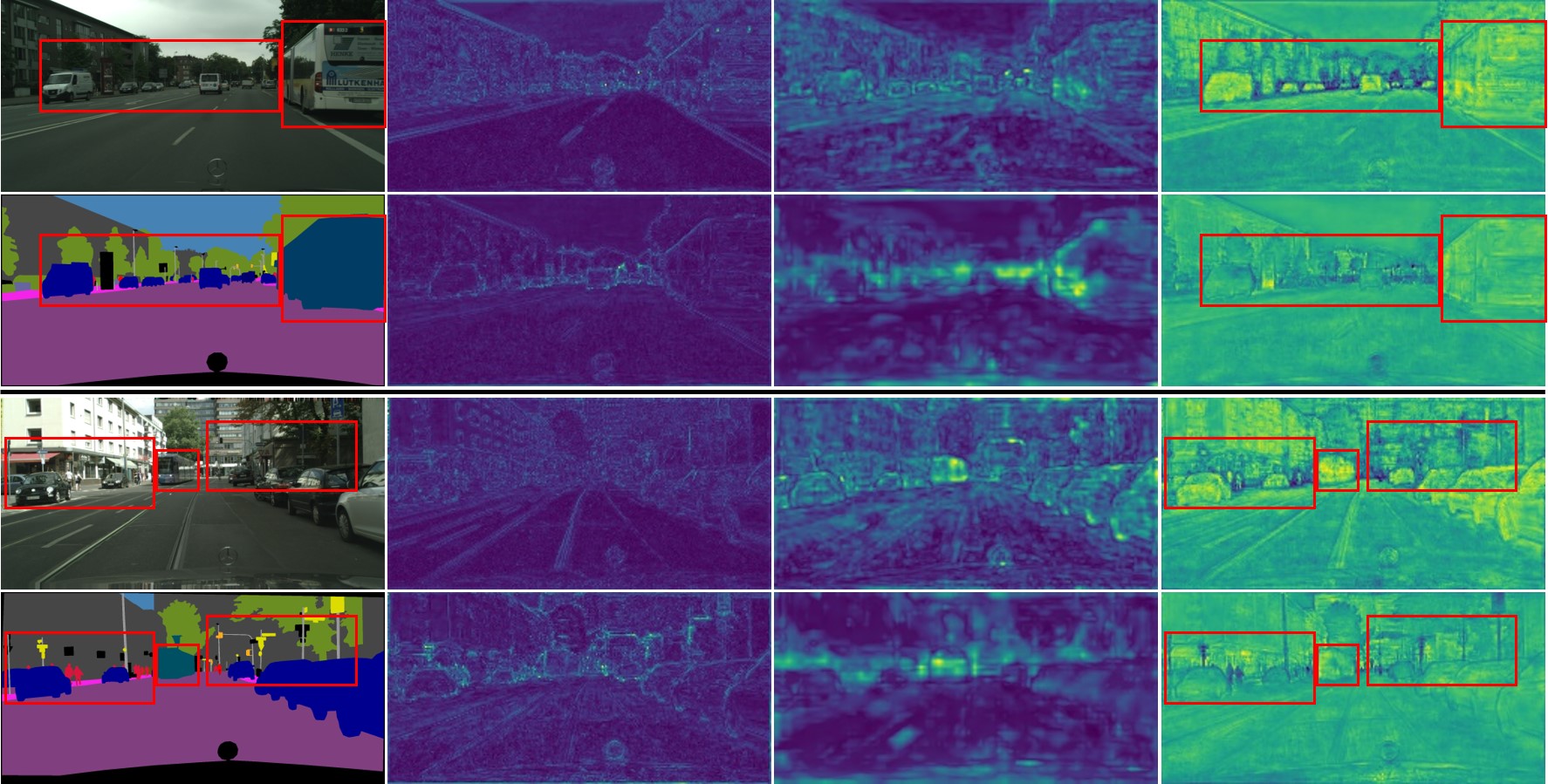}
\caption{Feature visualization of Pag module. The maps in the first row from left to right are the original input image, P input, I input and output of Sigmoid function for the first Pag; The maps in the second row are groudtruth, P, I inputs and Sigmoid output for the second Pag; The third and fourth rows are for another image.}
\label{fig:pag_map}
\end{figure}

\begin{figure}[t]
\centering
    \includegraphics[width=8.3cm]{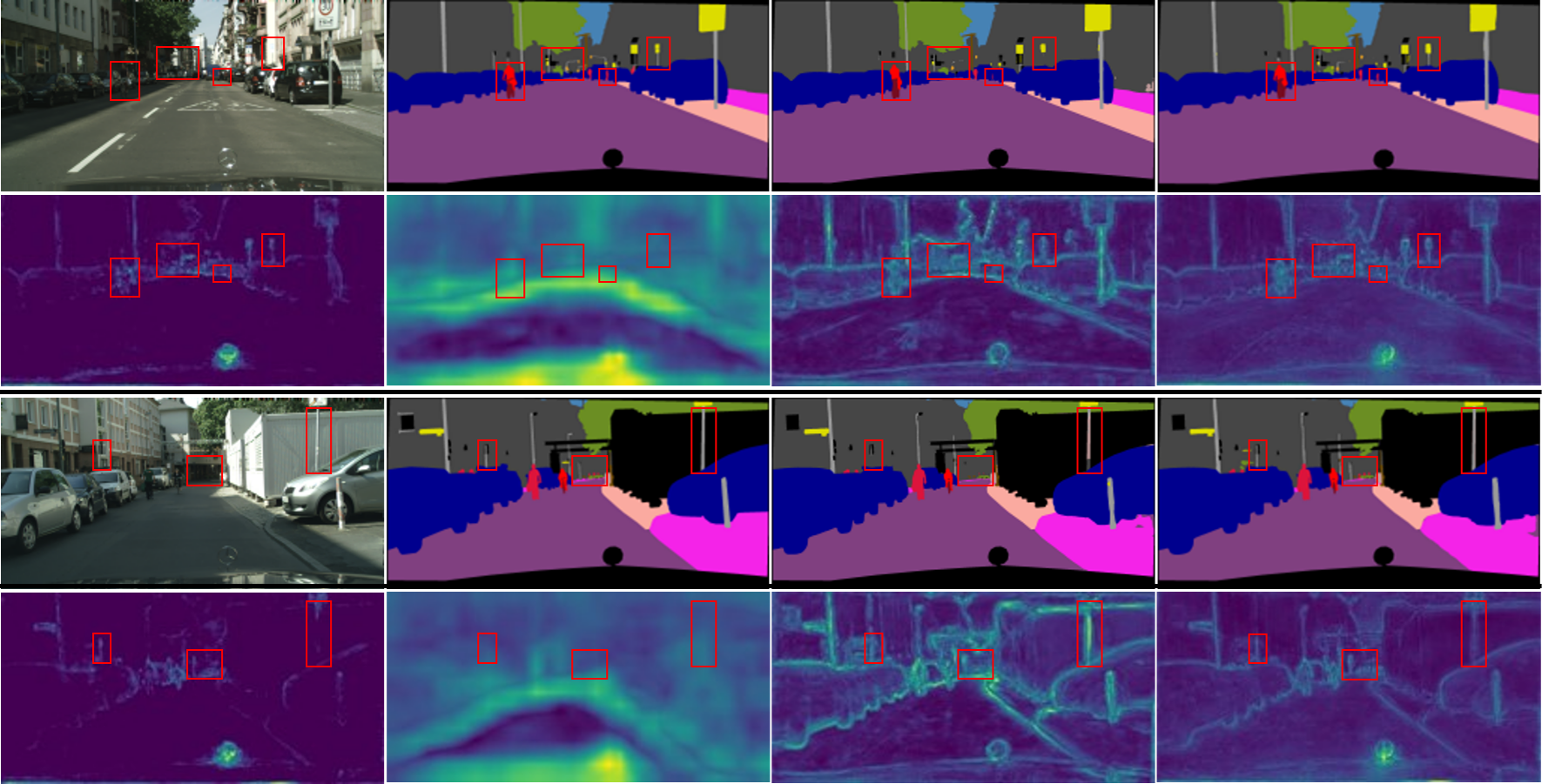}
\caption{Feature visualization of Bag module. The maps in the first row from left to right are the original input image, ground truth, predictions of DDRNet-23 and PIDNet-M; The maps in the second row are P, I and D input and final output for Light-Pag in PIDNet-M; The third and fourth rows are for another image.}
\label{fig:bag_map}
\end{figure}

\noindent
\textbf{Collaboration of Pag and Bag.} P branch utilizes Pag module to learn useful information from I branch without being overwhelmed before fusion stage and Bag module is introduced to guide the fusion of detailed and context features. As Table \ref{tab:pag_bag} shows, lateral connection could significantly improve the model accuracy and pretraining could further boost its performance.
\begin{table}[h]
\centering
\begin{tabular}{cccccccc} 
\Xhline{1pt}
\multirow{2}{*}{IM} & \multicolumn{3}{c}{Lateral} &  & \multicolumn{2}{c}{Fusion} & \multirow{2}{*}{mIOU}  \\ 
\cline{2-4}\cline{6-7}
                          & None & Add & Pag                       &  & Add & Bag                         &                        \\ 
\Xhline{1pt}
                          &      &\checkmark      &                         &  &\checkmark    &                             & 79.3                   \\ 
\hline
                          &      &     & \checkmark                         &  & \checkmark    &                            & 78.1                   \\ 
\hline
\checkmark                         & \checkmark    &     &                           &  &\checkmark   &                              & 80.0                  \\ 
\hline

\checkmark                         &      &\checkmark     &                          &  &\checkmark    &                             & 80.7                   \\ 
\hline
\checkmark                         &      &    & \checkmark                          &  & \checkmark    &                            & 80.5                   \\ 
\hline
\checkmark                         &      &\checkmark     &                          &  &    &\checkmark                             & 80.5                   \\ 
\hline
\checkmark                         &      &     & \checkmark                         &  &     & \checkmark                           & \textbf{80.9}                   \\
\Xhline{1pt}
\end{tabular}
\caption{Ablation study of Pag and Bag on PIDNet-L. IM refers to ImageNet \cite{imagenet} pretraining, Add represents the element-wise summation operation and None means there is no lateral connection.}
\label{tab:pag_bag}
\end{table}
In our scenario, the combinations of Add lateral connection and Bag fusion module or Pag lateral connection and Add fusion module make little sense since preservation of details should be consistent in the entire network. Thus, we only need to compare the performance of Add + Add and Pag + Bag and the experimental results in Table \ref{tab:pag_bag} and \ref{tab:pappm_bag} demonstrate the superiority of the collaboration of Pag and Bag (or Light-Bag). The visualization of feature maps in Figure \ref{fig:pag_map} shows that the small objects become much darker compared with large objects in the Sigmoid map for second Pag, where I branch loses more detailed information. Also, the features in boundary regions and small objects are greatly enhanced in the output of Bag module, which is illustrated in Figure \ref{fig:bag_map} and explains the reason why we choose coarse boundary detection. 

\label{exp}

\begin{table}[h]
\centering
\begin{tabular}{ccccccc} 
\Xhline{1pt}
\multicolumn{2}{c}{PPM} &  & \multicolumn{2}{c}{Fusion} & \multirow{2}{*}{mIOU} & \multirow{2}{*}{FPS}  \\ 
\cline{1-2}\cline{4-5}
DAPPM & PAPPM           &  & Add & Bag                  &                       &                       \\ 
\Xhline{1pt}
\checkmark     &                 &  &     & \checkmark                    & \textbf{78.8}                  & 83.7                  \\ 
\hline
      & \checkmark               &  & \checkmark   &                      & 78.4                  & \textbf{97.8}                  \\ 
\hline
      & \checkmark               &  &     & \checkmark                    & \textbf{78.8}                  & 93.2                  \\
\Xhline{1pt}
\end{tabular}
\caption{Ablation study of PAPPM and Light-Bag on PIDNet-S.}
\label{tab:pappm_bag}
\end{table}
\noindent
\textbf{Efficiency of PAPPM.} For real-time models, a heavy context aggregation module could drastically slow down the inference speed and may surpass the representation ability of the network. Thus, we proposed the PAPPM, which is constituted by parallel structure and small number of parameters. The experimental results in Table \ref{tab:pappm_bag} show that PAPPM achieves the same accuracy as DAPPM \cite{ddrnet} but presents a speed-up of 9.5 FPS for our light-weight model.

\begin{table}[h]
\centering
\begin{tabular}{ccccc} 
\Xhline{1pt}
\multicolumn{3}{c}{Extra Loss}                                        & \multirow{2}{*}{OHEM} & \multirow{2}{*}{mIOU}     \\ 
\cline{1-3}
$l_{0}$                    & $l_{1}$                     & $l_{3}$                     &                       &                           \\ 
\Xhline{1pt}
                      &                       &                       &                       & 78.6                      \\ 
\hline
\checkmark                      &                       &                       &                       & 78.8                     \\ 
\hline
\checkmark                      & \checkmark                      &                       &                       & 79.9                      \\ 
\hline
\checkmark                     & \checkmark                      & \checkmark                      &                       & 80.5                      \\ 
\hline
\multicolumn{1}{c}{\checkmark} & \multicolumn{1}{c}{\checkmark} & \multicolumn{1}{c}{\checkmark} & \checkmark                      & \multicolumn{1}{c}{\textbf{80.9}}  \\
\Xhline{1pt}
\end{tabular}
\caption{Ablation study of extra losses and OHEM for PIDNet-L.}
\label{tab:loss}
\end{table}
\noindent
\textbf{Effectiveness of Extra losses.} Three extra losses were introduced to PIDNet to boost the optimization of entire network and emphasize the functionality for each components. According to Table \ref{tab:loss}, boundary loss $l_{1}$ and boundary-awareness loss $l_{3}$ are necessary for better performance, especially the boundary loss (+1.1\% mIOU), which strongly proves the necessity of D branch, and Online Hard Example Mining (OHEM) \cite{ohem} further improves the accuracy.

\subsection{Comparison}
\noindent
\textbf{CamVid.} For CamVid \cite{camvid} dataset, only the accuracy of DDRNet is comparable with our models, so we test its speed on our platform with the same setting for fair comparison considering our platform is more advanced than theirs. 
\begin{table}[h]
\centering
\begin{tabular}{lccc} 
\Xhline{1pt}
Model           & mIOU & \#FPS & GPU         \\ 
\Xhline{1pt}
MSFNet \cite{msfnet}          & 75.4 & 91.0  & GTX 2080Ti  \\ 
\hline
PP-LiteSeg-T \cite{ppseg}    & 75.0 & 154.8 & GTX 1080Ti  \\ 
\hline
TD2-PSP50 \cite{td_psp}       & 76.0 & 11.0  & TITAN X     \\ 
\hline
BiSeNetV2\textsuperscript{\textdagger} \cite{bisenetv2}     & 76.7 & 124.0 & GTX 1080Ti  \\
BiSeNetV2-L\textsuperscript{\textdagger} \cite{bisenetv2}   & 78.5 & 33.0  & GTX 1080Ti  \\ 
\hline
HyperSeg-S \cite{hyperseg}    & 78.4 & 38.0  & GTX 1080Ti  \\
HyperSeg-L \cite{hyperseg}      & 79.1 & 16.6  & GTX 1080Ti  \\ 
\hline
DDRNet-23-S\textsuperscript{\textdagger}\textsuperscript{\text{*}} \cite{ddrnet}& 78.6 & \textbf{182.4} & RTX 3090    \\
DDRNet-23\textsuperscript{\textdagger}\textsuperscript{\text{*}} \cite{ddrnet}      & 80.6 & 116.8 & RTX 3090    \\ 
\hline
PIDNet-S\textsuperscript{\textdagger}        & 80.1 & 153.7 & RTX 3090    \\
PIDNet-S-Wider\textsuperscript{\textdagger}        & \textbf{82.0} & 85.6  & RTX 3090    \\
\Xhline{1pt}
\end{tabular}
\caption{Comparison of speed and accuracy on CamVid. The models pretrained by Cityscapes \cite{cityscapes} are marked with \textdagger; The inference speeds for models marked with \text{*} are tested on our platform.}
\label{tab:camvid}
\end{table}
The experimental results in Table \ref{tab:camvid} show that the accuracy of all our models exceeds 80\% mIOU and PIDNet-S-Wider, which simply doubles the number of channels for PIDNet-S, achieves the highest accuracy with a big margin ahead of previous models. Besides, the accuracy of PIDNet-S surpasses previous state-of-art model: DDRNet-23-S by 1.5\% mIOU with only around 1 ms latency increase. 

\begin{table*}[t]
\centering
\begin{tabular}{lC{0.8cm}C{0.8cm}C{1.0cm}cccc} 
\Xhline{1pt}
\multirow{2}{*}{Model} & \multicolumn{2}{c}{mIOU} & \multirow{2}{*}{\#FPS} & \multirow{2}{*}{GPU} & \multirow{2}{*}{Resolution} & \multirow{2}{*}{\#GFLOPs} & \multirow{2}{*}{\#Params}  \\ 
\cline{2-3}
                       & Val  & Test              &                        &                      &                             &                           &                            \\ 
\Xhline{1pt}
MSFNet  \cite{msfnet}               & -   & 77.1              & 41                     & RTX 2080Ti           & 2048$\times$1024                   & 96.8                      & -                         \\ 
\hline
DF2-Seg1 \cite{df_seg}              & 75.9 & 74.8              & 67.2                   & GTX 1080Ti           & 1536$\times$768                    & -                        & -                         \\
DF2-Seg2 \cite{df_seg}              & 76.9 & 75.3              & 56.3                   & GTX 1080Ti           & 1536$\times$768                    & -                        & -                         \\ 
\hline
SwiftNetRN-18 \cite{swiftnet}         & 75.5 & 75.4              & 39.9                   & GTX 1080Ti           & 2048$\times$1024                   & 104.0                     & 11.8M                      \\
SwiftNetRN-18 ens \cite{swiftnet}     & -   & 76.5              & 18.4                   & GTX 1080Ti           & 2048$\times$1024                   & 218.0                     & 24.7M                      \\ 
\hline
CABiNet \cite{cabinet}               & 76.6 & 75.9              & 76.5                   & RTX 2080Ti           & 2048$\times$1024                   & 12.0                      & 2.64M                      \\ 
\hline
BiSeNet(Res18) \cite{bisenet}        & 74.8 & 74.7              & 65.5                   & GTX 1080Ti           & 1536$\times$768                    & 55.3                      & 49M                        \\
BiSeNetV2-L  \cite{bisenetv2}          & 75.8 & 75.3              & 47.3                   & GTX 1080Ti           & 1024$\times$512                    & 118.5                     & -                         \\ 
\hline
STDC1-Seg75\textsuperscript{\text{*}} \cite{stdc}           & 74.5 & 75.3              & 74.8                  & RTX 3090           & 1536$\times$768                    & -                        & -                         \\
STDC2-Seg75\textsuperscript{\text{*}} \cite{stdc}           & 77.0 & 76.8              & 58.2                   & RTX 3090           & 1536$\times$768                    & -                        & -                         \\ 
\hline
PP-LiteSeg-T2\textsuperscript{\text{*}} \cite{ppseg}         & 76.0 & 74.9              & 96.0                  & RTX 3090           & 1536$\times$768                    & -                        & -                         \\
PP-LiteSeg-B2\textsuperscript{\text{*}} \cite{ppseg}         & 78.2 & 77.5              & 68.2                  & RTX 3090           & 1536$\times$768                    & -                        & -                         \\ 
\hline
HyperSeg-M\textsuperscript{\text{*}} \cite{hyperseg}             & 76.2 & 75.8              & 59.1                   & RTX 3090           & 1024$\times$512                    & 7.5                       & 10.1                       \\
HyperSeg-S\textsuperscript{\text{*}} \cite{hyperseg}             & 78.2 & 78.1              & 45.7                   & RTX 3090           & 1536$\times$768                    & 17.0                      & 10.2                       \\ 
\hline
SFNet(DF2)\textsuperscript{\text{*}}   \cite{sfnet}           & -   & 77.8              & 87.6                   & RTX 3090             & 2048$\times$1024                   & -                        & 10.53M                     \\
SFNet(ResNet-18)\textsuperscript{\text{*}}  \cite{sfnet}       & -   & 78.9              & 30.4                   & RTX 3090             & 2048$\times$1024                   & 247.0                     & 12.87M                     \\
SFNet(ResNet-18)\textsuperscript{\textdagger}\textsuperscript{\text{*}}  \cite{sfnet}      & -   & 80.4              & 30.4                   & RTX 3090             & 2048$\times$1024                   & 247.0                     & 12.87M                     \\ 
\hline
DDRNet-23-S\textsuperscript{\text{*}}  \cite{ddrnet}        & 77.8 & 77.4              & \textbf{108.1}                  & RTX 3090             & 2048$\times$1024                   & 36.3                      & 5.7M                       \\
DDRNet-23\textsuperscript{\text{*}}  \cite{ddrnet}              & 79.5 & 79.4              & 51.4                   & RTX 3090             & 2048$\times$1024                   & 143.1                     & 20.1M                      \\
DDRNet-39\textsuperscript{\text{*}}  \cite{ddrnet}              & -   & 80.4              & 30.8                   & RTX 3090             & 2048$\times$1024                   & 281.2                     & 32.3M                      \\ 
\hline
\rowcolor{green!20} PIDNet-S-Simple               & 78.8 & 78.2              & 100.8                   & RTX 3090             & 2048$\times$1024                   & 46.3                      & 7.6M                       \\
\rowcolor{green!20} PIDNet-S               & 78.8 & 78.6              & 93.2                   & RTX 3090             & 2048$\times$1024                   & 47.6                      & 7.6M                       \\
\rowcolor{cyan!20} PIDNet-M               & \textbf{80.1} & 80.1              & 39.8                   & RTX 3090             & 2048$\times$1024                   & 197.4                     & 34.4M                      \\
\rowcolor{cyan!20} PIDNet-L               & \textbf{80.9} & \textbf{80.6}              & 31.1                   & RTX 3090             & 2048$\times$1024                   & 275.8                     & 36.9M                      \\
\Xhline{1pt}
\end{tabular}
\caption{Comparison of speed and accuracy on Cityscapes. The models pretrained by other segmentation datasets are marked with \textdagger; The inference speeds for models marked with \text{*} are tested on our platform. The GFLOPs for PIDNet is derived based on full-resolution input.}
\label{tab:cityscapes}
\end{table*}

\noindent
\textbf{Cityscapes.} Previous real-time works treat Cityscapes \cite{cityscapes} as the standard benchmark considering its high-quality annotation. As shown in Table \ref{tab:cityscapes}, we test the inference speeds of the models published in recent two years on the same platform without any acceleration tool as PIDNets for fair comparison. The experimental results show that PIDNets achieve the best trade-off between inference speed and accuracy. Specifically, PIDNet-L surpasses SFNet(ResNet-18)\textsuperscript{\textdagger} and DDRNet-39 in terms of speed and accuracy and becomes the most accurate model in real-time domain by rising the test accuracy from 80.4\% to 80.64\% mIOU. PIDNet-M and PIDNet-S also provide much higher accuracy compared with other models with similar inference speeds. Removing Pag and Bag modules from PIDNet-S, we provide an even faster option: PIDNet-S-Simple, which has weaker generalization ability but still presents highest accuracy among models with latency less than 10 ms.

\noindent
\textbf{PASCAL Context.} The $Avg (17,8)$ path in PAPPM is removed since the image size is too small in PASCAL Context \cite{pascal_context}. Different from other two datasets, multi-scale and flip inference are utilized here for fair comparison with previous models. Even though there are less detailed annotations in PASCAL Context compared with previous two datasets, our models still achieve competitive performance among existing heavy networks, as shown in Table \ref{tab:pascal}.  
\begin{table}[ht]
\centering
\begin{tabular}{lccc} 
\Xhline{1pt}
Model            & BaseNet & mIOU-59  & mIOU-60    \\ 
\Xhline{1pt}
DeepLab-v2\cite{deeplabv2}   & D-Res-101 & -  & 45.7   \\ 
\hline
RefineNet \cite{refinenet}   & Res-152 & -  & 47.3   \\ 
\hline
PSPNet  \cite{pspnet}       & D-Res-101 & 47.8  & -   \\ 
\hline
Ding et al.  \cite{ding2018context}    & D-Res-101  & 51.6 & - \\ 
\hline
EncNet  \cite{encnet}    & D-Res-101  & 52.6 & - \\ 
\hline
HRNet \cite{hrnet}    & V2-W48  & 54.0 & 48.3 \\ 
\hline
PIDNet-M        & -  & 51.0  & 46.0 \\
PIDNet-L        & -   & 51.9 & 46.6    \\
\Xhline{1pt}
\end{tabular}
\caption{Comparison of accuracy on Pascal-Context (w/ and w/o background class). D-Res-101 refers to Dilated ResNet-101.}
\label{tab:pascal}
\end{table}

\section{Conclusion}
\label{sec:conclusion}
This paper presents a novel three-branch network architecture: PIDNet for real-time semantic segmentation. PIDNet achieves the best trade-off between inference time and accuracy. However, since PIDNet utilizes the boundary prediction to balance the detailed and context information, precise annotation around boundary, which usually requires a large amount of time, is preferred for better performance.


{\small
\bibliographystyle{ieee_fullname}
\bibliography{egbib}
}

\end{document}